%% file: main.tex
\documentclass[10pt,twocolumn,letterpaper]{article}

\usepackage{cvpr}              
\usepackage{amsmath,amssymb}
\usepackage{amsthm}
\usepackage{xspace}
\usepackage[dvipsnames, table]{xcolor}
\usepackage{tcolorbox}
\usepackage{float}
\theoremstyle{plain}
\newtheorem{proposition}{Proposition}
\newcommand{\method}{{\tt{NewtonRewards}}\xspace}

\input{preamble}
\definecolor{cvprblue}{rgb}{0.7,0.11,0.11}
\usepackage[pagebackref,breaklinks,colorlinks,allcolors=cvprblue]{hyperref}

\def\paperID{6729} 
\def\confName{CVPR}
\def\confYear{2026}

\title{What about gravity in video generation? 
\\Post-Training Newton's Laws with Verifiable Rewards 
} 

\author{Minh-Quan Le$^{1}$ \quad
Yuanzhi Zhu$^{2}$ \quad
Vicky Kalogeiton$^{2,\dagger}$ \quad
Dimitris Samaras$^{1,\dagger}$\\
$^1$Stony Brook University \quad $^2$LIX, École Polytechnique, CNRS, IPP \quad
$^\dagger$Equal Advising\\
\href{https://cvlab-stonybrook.github.io/NewtonRewards}{https://cvlab-stonybrook.github.io/NewtonRewards}
}

\begin{document}
\maketitle

\input{sec/0_abstract}    
\input{sec/1_intro}
\input{sec/2_related_works}
\input{sec/3a_newton}

\input{sec/3_method_new}

\input{sec/4_dataset}
\input{sec/5_experiment}

\input{sec/6_conclusion}

\clearpage

\section*{Acknowledgements} 
This research was supported by NSF grant IIS-2212046, ANR-22-CE23-0007, ANR-22-CE39-0016, Hi!Paris grant and fellowship, DATAIA Convergence Institute as part of the ``Programme d’Investissement d'Avenir'' (ANR-17-CONV-0003) operated by Ecole Polytechnique, IP Paris, and was granted access to the IDRIS High-Performance Computing (HPC) resources under the allocation 2025-AD011015894 made by GENCI. 

{
    \small
    \bibliographystyle{ieeenat_fullname}
    \bibliography{main}
}

\clearpage
\appendix
\input{sec/X_suppl}

\end{document}

%% file: preamble.tex
\usepackage{tikz}
\usepackage{booktabs}
\usepackage{multirow}
\usepackage{xcolor}



%% file: sec/0_abstract.tex
\begin{abstract} 
Recent video diffusion models can synthesize visually compelling clips, yet often violate basic physical laws--objects float, accelerations drift, and collisions behave inconsistently--revealing a persistent gap between visual realism and physical realism. We propose \method, the first physics-grounded post-training framework for video generation based on \emph{verifiable rewards}. Instead of relying on human or VLM feedback, \method extracts \emph{measurable proxies} from generated videos using frozen utility models: optical flow serves as a proxy for velocity, while high-level appearance features serve as a proxy for mass. These proxies enable explicit enforcement of Newtonian structure through two complementary rewards: a Newtonian kinematic constraint enforcing constant-acceleration dynamics, and a mass conservation reward preventing trivial, degenerate solutions. 
We evaluate \method on five Newtonian Motion Primitives (free fall, horizontal/parabolic throw, and ramp sliding down/up) using our newly constructed large-scale benchmark, \texttt{NewtonBench-60K}. Across all primitives in visual and physics metrics, \method consistently improves physical plausibility, motion smoothness, and temporal coherence over prior post-training methods. It further maintains strong performance under out-of-distribution shifts in height, speed, and friction. Our results show that physics-grounded verifiable rewards offer a scalable path toward physics-aware video generation. 
\end{abstract}

%% file: sec/1_intro.tex
\section{Introduction}
\label{sec:intro}

Gravity is everywhere. From the fall of an apple to the motion of celestial bodies, 
physical laws 
govern how objects move, interact, and persist through time. Yet, in the rapidly advancing field of video generation, such fundamental principles are largely absent \cite{Motamed_2025_PhysicalPrinciples,bansal2025videophy,Kang_2025_Farvideogenerationworld,Kong_2024_Hunyuanvideo,zhang2025morpheus,liu2025generative}. Recent generative models can produce visually stunning videos from text \cite{zheng2024open,yang2024cogvideox,hacohen2024ltx,gao2025seedance,wan2025}, images \cite{svdblattmann2023,hacohen2024ltx,gao2025seedance,wan2025}, or latent trajectories~\cite{courant2024et}, but 
often exist in worlds unbound by physics: worlds where objects float \cite{yuan2025likephys}, collisions resolve unrealistically \cite{bansal2025videophy}, and motion unfolds without cause \cite{zhang2025morpheus}. These violations of basic laws, such as Newton’s laws of motion, highlight a lack of \textit{physical realism} in video generation.

\begin{figure}[t!]
    \centering
    \includegraphics[width=\linewidth]{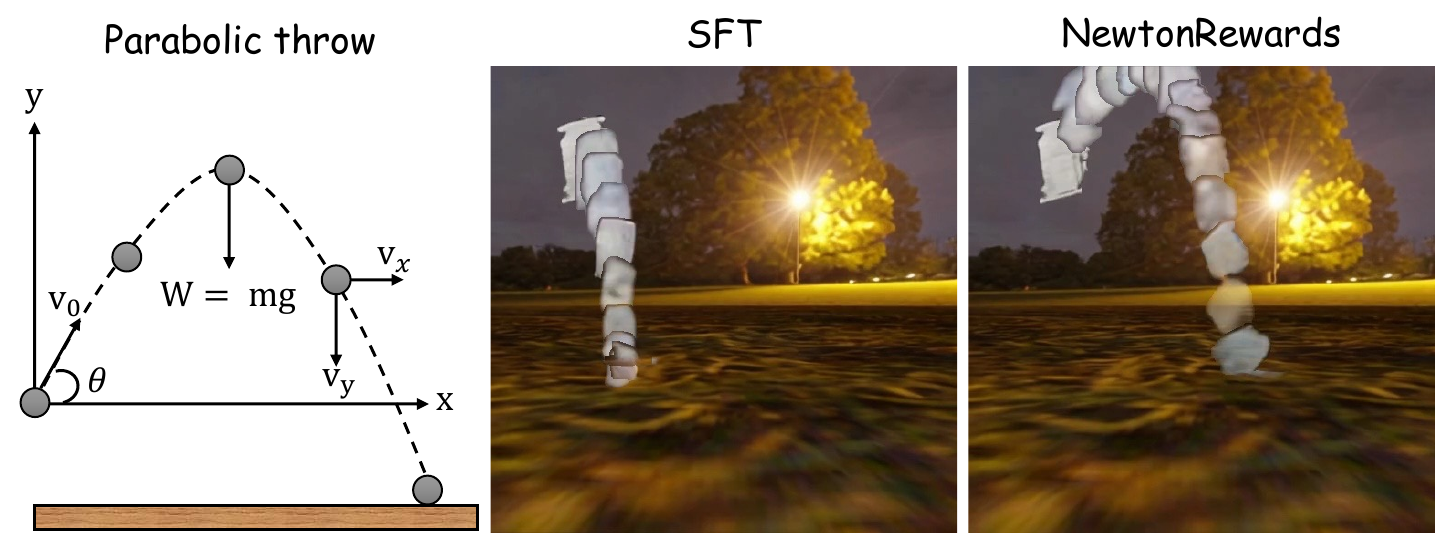}
    \vspace{-6mm}
    \caption{\method enforce physical laws in video generation. Shown is a parabolic throw scenario from our NewtonBench-60K dataset. Baseline supervised fine-tuning (SFT) produces implausible motion violating Newtonian dynamics. Our \method post-training restores parabolic trajectories that follow constant-acceleration behavior predicted by physics.}
    \label{fig:teaser}
    \vspace{-10pt}
\end{figure}

Embedding physical plausibility into video generation is more than an aesthetic choice; it is a necessity.  In many applications, from 
immersive game environments and realistic cinematic worlds~\cite{wang2025akira}, to training world models for games \cite{bruce2024genie,kanervisto2025world,yu2025gamefactory}, autonomous driving \cite{hu2023gaia,russell2025gaia,agarwal2025cosmos,feng2025survey} and robotic control \cite{yang2023learning,bar2025navigation,li2025robotic}, generated videos serve as data for perception, reasoning, and action \cite{wu2025rlvr}. In these scenarios, non-physical dynamics can lead to inconsistent learning signals, unrealistic affordances, and failure to generalize to the real world. A model that understands that ``objects fall down'' or that ``collisions change velocity'' produces not just more believable motion, but also a better 
world model.

Recently, several approaches have sought to embed physical plausibility into video generation.
These range from methods that fine-tune diffusion models using textual or feedback-based supervision from large language or vision-language models~\cite{wang2025wisa,zhang2025think,hao2025enhancing,yang2025vlipp,garrido2025intuitive,zhang2025videorepa,motamed2025travl,Xue_2025_PhyT2V,le2025hummingbird}, 
to approaches that incorporate physical simulators or 3D representations as motion or geometry priors~\cite{xie2025physanimator,wang2025physctrl,zhao2025synthetic,li2025pisa,liu2024physgen,yuan2025newtongen,Lv_2024_Gpt4motion,Hsu_2024_Autovfx}, and 
to those that rely on physics-rich datasets or post-training signals derived from real or synthetic videos~\cite{li2025pisa,wang2025wisa,chen2025hierarchical,ji2025physmaster}.
Still, these methods typically use physics signals coming from humans or VLM feedback as a condition rather than explicitly enforcing physics laws.

Neither humans nor ``VLMs-as-judge" can precisely evaluate how well physical constraints are being followed (apart from egregious physical law violations). 
As a result, generated videos often appear visually realistic yet fail to satisfy physics principles such as momentum conservation, force–acceleration proportionality, and consistent gravitational effects.
We argue that despite improved perceptual realism and motion smoothness in recent work, 
 physically plausible video generation remains a challenge, as models must respect the physical laws governing object dynamics.

Therefore, in this work, we propose the first verifiable rewards for physical laws in video generation, which employ rule-based evaluations that can automatically verify the correctness of the output \cite{lambert2024tulu,wei2025asymmetry,shao2024deepseekmath,guo2025deepseek,mroueh2025reinforcement,wen2025reinforcement}. 
Given a video generator, physical quantities such as velocity or force cannot be directly observed from its raw output frames.
To bridge this gap, we estimate these quantities using pre-trained utility models (e.g., optical flow or video-embedding networks).
Their outputs, which we term {\em measurable proxies}, serve as observable surrogates for underlying physical variables.
By defining physics constraints and rewards on these proxies, we post-train video generators to produce videos that follow physical laws explicitly.

As an initial application of this approach, we present \method{}, as a basis for a framework for Newtonian motion constraints. 
We define measurable proxies (optical flow and appearance embeddings) for physical constraints (velocity and mass) and use them to formulate both kinematic and mass-conservation rewards. 
We evaluate our framework across five Newtonian Motion Primitives (NMPs): 
(i-iii) \emph{free fall, horizontal and parabolic throw}, and (iv-v) \emph{ramp sliding down/up}. 
We experiment on our large-scale simulated dataset, \textbf{NewtonBench-60K}, specifically designed for dynamic motion evaluation, with diverse scenarios for each NMP.
Empirically, \method{} consistently improves physical plausibility, motion smoothness, and temporal coherence over prior methods such as PISA~\cite{li2025pisa}. It yields gains across all five NMPs for both in-distribution (ID) and out-of-distribution (OOD) settings. 
Physics-grounded constraints correct kinematic violations that visual feature alignment alone cannot fix, i.e., reducing constant-acceleration residuals and mitigating reward hacking behaviors (objects vanish to minimize motion).

Our contributions can be summarized as:
\begin{itemize}
    \item We introduce \method{}, an elegant physics-grounded post-training framework for video generation that explicitly enforces Newtonian dynamic motions (e.g., throw, free fall, slide with friction).
    \item We employ optical flow and visual features as a differentiable proxy to devise Newtonian kinematic and mass constraints, yielding verifiable, rule-based rewards that promote physically correct motion.
   \item We simulate a controlled, large-scale dataset and benchmark specifically designed to evaluate dynamic motion realism and physical consistency in video generation. We will release our simulation/training code, \texttt{NewtonBench-60K}, and models to the community.
   \item Experiments show \method consistently improves across both visual and physics metrics, across all five NMPs, for both ID and OOD settings, producing more physically faithful and temporally coherent videos.
    
\end{itemize}

We posit that our methodology is general. \emph{Given a measurable proxy of a variable in a physical law, the same steps can be followed to come up with verifiable rewards appropriate for that law.} We hope that this general framework can pave the way for future research in this area.

%% file: sec/2_related_works.tex
\section{Related Works}







\noindent \textbf{Video Generative Models.}
Video models are progressing rapidly \cite{singer2022make,Ho_2022_VDM,imagenvideo2022,svdblattmann2023,emuvideo2023,bar2024lumiere,sora2024,blattmann2023align,wang2025lavie,yu2023language}. Using large-scale datasets and scalable backbones (e.g., DiT-style architectures \cite{Peebles_2023_DiT}), modern models produce photorealistic short clips conditioned on text, images, and other control signals \cite{lee2022sound,wang2025akira,ma2025controllable,peng2024controlnext,zhang2024mimicmotion,zhang2024moonshot,guo2024sparsectrl,bhowmik2025moalign}. 
Open-source efforts, such as OpenSora~\cite{zheng2024open}, CogVideo~\cite{Hong_2023_Cogvideo,yang2024cogvideox}, HunyuanVideo~\cite{Kong_2024_Hunyuanvideo}, and Wan 2.1~\cite{wan2025}, have 
shown big improvements in visual fidelity and conditional control.
Although these models show some emergent reasoning on tasks beyond their training \cite{wiedemer2025video},  scaling data or model size alone often cannot eliminate unrealistic motion or physics violations \cite{Kang_2025_Farvideogenerationworld,Motamed_2025_PhysicalPrinciples}.

\noindent \textbf{Physics-aware Video Generation.}
There are 3 groups of strategies for 
physics priors/dynamics in video generation. 

\textit{(1) Instruction and feedback-based fine-tuning.}
Several methods fine-tune video diffusion models using feedback or textual instructions from Large Language Models (LLMs) \cite{lian2023llm,zhang2025think,hao2025enhancing} or Vision-Language Models (VLMs) \cite{yang2025vlipp,garrido2025intuitive,zhang2025videorepa,motamed2025travl}. 
For instance, PhyT2V \cite{Xue_2025_PhyT2V} introduces a feedback loop where an LLM checks if generated videos obey physical laws, reasoning over captions from a video captioning model.
However, such feedback is indirect and often reflects perception rather than physics consistency.


\textit{(2) Physics-guided simulation and representation.}
Another line of work leverages physics simulators or 3D representations to guide video generation.
Some approaches pre-compute 3D or physically plausible conditions as auxiliary input for the video generator~\cite{xie2025physanimator,wang2025physctrl,yuan2025newtongen}. 
Others build integrated pipelines combining simulation, rendering, and generative modeling to ensure motion realism~\cite{Lv_2024_Gpt4motion,liu2024physgen,Hsu_2024_Autovfx,Xue_2025_PhyT2V}. 
For instance, PhysGen~\cite{liu2024physgen} 
performs image-based warping with simulated motion dynamics given user-defined forces and torques.
However, relying on external simulators enforces physics only indirectly, through the pre-trained model’s interpretation of conditioned data.

\textit{(3) Physics-rich datasets and post-training.}
This strategy focuses on data-level physical grounding.
PISA \cite{li2025pisa} introduced a dataset of 361 real-world and 60 synthetic videos of objects falling in diverse environments and used post-training strategies such as supervised fine-tuning and object-reward optimization, aligning optical flow, depth, and segmentation maps to improve physical consistency.
Methods such as \cite{zhao2025synthetic,wang2025wisa,chen2025hierarchical} rely on physics-rich or instruction-grounded datasets to enhance physical plausibility in a data-driven manner.
Similarly, PhysMaster \cite{ji2025physmaster} proposed optimizing a neural \textit{PhysEncoder} via reinforcement learning, using costly human-annotated preference data.


%% file: sec/3a_newton.tex
\section{Newtonian Motion Primitives (NMPs)}

We formulate post-training of video diffusion models through classical mechanics, grounded in Newton’s three laws of motion (Section~\ref{sub:backgroundnewton}). In our setting, an object observed in a video sequence undergoes motion determined by external forces acting upon it.
Building upon these physical laws, we identify five canonical \emph{Newtonian Motion Primitives (NMPs)}: \emph{free fall}, \emph{horizontal throw}, \emph{parabolic throw}, \emph{ramp sliding down}, and \emph{ramp sliding up}.  
Each primitive corresponds to a distinct combination of forces and produces a characteristic pattern of constant acceleration in the image plane, as
described in detail in Section~\ref{sec:NMPs}.

\begin{figure}[t!]
    \centering
    \includegraphics[width=\linewidth]{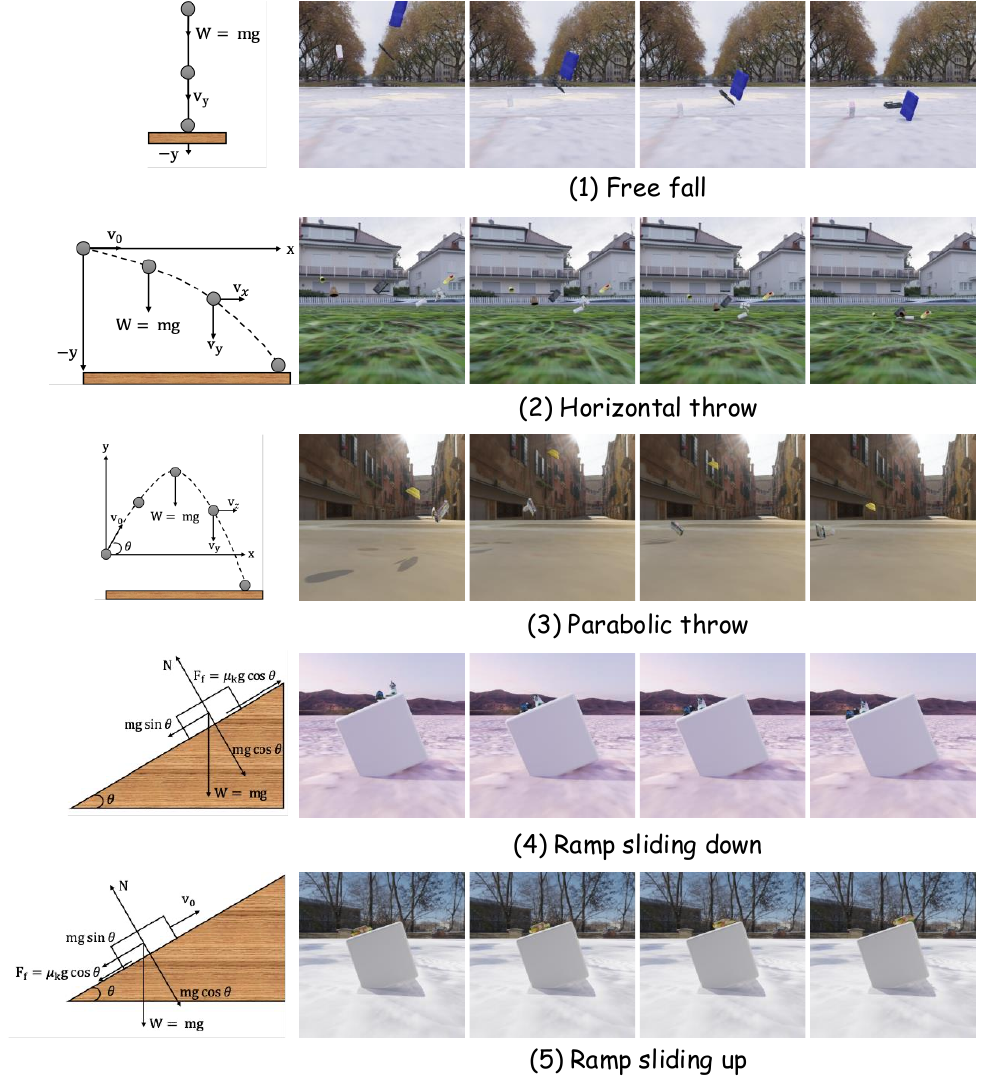}
    \vspace{-15pt}
\caption{Illustration of the five NMPs in the proposed \texttt{NewtonBench-60K} dataset. Left: corresponding free-body diagrams showing dominant forces and accelerations. Right: rendered trajectories from our Kubric-based \cite{greff2022kubric} simulator, demonstrating constant-acceleration dynamics in diverse environments.}    
    \label{fig:dataset}
    \vspace{-10pt}
\end{figure}

\subsection{Background: Newton's Laws of Motion}
\label{sub:backgroundnewton}

Let $\mathbf{F}_{\text{net}}$ denote the net resultant force acting on the object, $\mathbf{v}=(\mathbf{v}_x,\mathbf{v}_y)$ its velocity projected onto the image plane, and $\mathbf{a}=\dot{\mathbf{v}}$ its corresponding acceleration. 

\noindent \textbf{Newton's First Law (Law of Inertia)} states that an object remains at rest or continues in uniform motion unless acted upon by an external force.  
It underlies scenarios such as \emph{free fall}, \emph{horizontal throw}, and \emph{parabolic throw}, where the absence of horizontal forces yields $\mathbf{a}_x = 0$ and constant $\mathbf{v}_x$.  

\noindent \textbf{Newton's Second Law (Law of Acceleration)} relates the net force $\mathbf{F}_{\text{net}}$ to the resulting acceleration $\mathbf{a}$ and mass $m$:
\begin{equation}
\mathbf{a} = \mathbf{F}_{\text{net}}/{{m}} \quad. 
\label{eq:newton2}
\end{equation}
It provides the quantitative foundation for our kinematic rewards, linking motion to underlying forces and mass.  

\noindent \textbf{Newton's Third Law (Action-Reaction)} states that when two bodies interact, they exert equal and opposite forces on each other.  
In our context, when an object interacts with a ramp or surface, the ramp exerts an equal and opposite normal force that balances the contact, together with a tangential frictional force opposing motion. 

\begin{figure*}[t!]
    \vspace{-15pt}
    \centering
    \includegraphics[width=\linewidth]{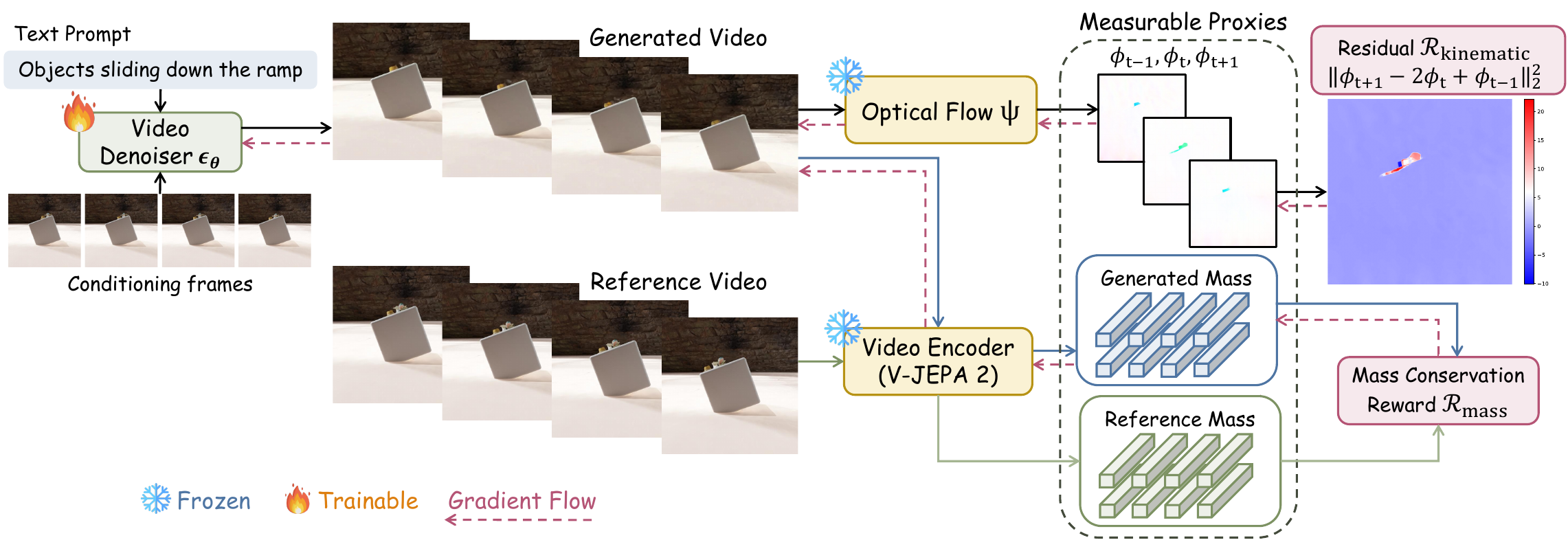}
    \vspace{-20pt}
\caption{\textbf{Physics-Grounded Video Post-Training Pipeline.} Our method improves a pre-trained video generator by using physics-based rewards. Utility models (optical flow $\Psi$ and V-JEPA 2) process the generated video to compute measurable proxies, from which kinematic and mass conservation rewards are derived to enforce explicit physics constraints.}  
    \vspace{-6pt}
    \label{fig:method}
    \vspace{-8pt}
\end{figure*}

\subsection{Forces and Accelerations of NMPs}
\label{sec:NMPs}


Let the world coordinate system be $(X,Y,Z)$, with gravity acting as $\mathbf{g} = (0, -g, 0)$, and let $(x,y)$ denote the projection in the image plane. 
Under a pinhole camera model with focal length $f$ and scene depth $Z$, the image-plane coordinates scale with depth, giving the approximation $x=(f/Z)X$ with proportional factor $f/Z$.
For short intervals, this scale is nearly constant, allowing analysis in image-space. 
\footnote{We assume that the image’s vertical axis is aligned with the direction of gravity.
For tilted cameras, projecting $\mathbf{g}$ onto the image plane yields constant apparent acceleration under mild assumptions: namely, weak perspective (small depth variation) and a static camera. 
}

\noindent\textbf{(1–3) NMP-F/TH/TP: Free Fall, Horizontal and Parabolic Throw.}
Object motion under uniform gravity $\mathbf{g}$ and no other external forces follows Newton’s Second Law:
\begin{equation}
\mathbf{a}_x = 0\quad ,\qquad \mathbf{a}_y = -g \quad,
\label{eq:freefall}
\end{equation}
corresponding to constant downward acceleration.
Different initial conditions yield special cases: zero initial velocity for free fall (NMP-F), nonzero horizontal velocity for horizontal throw (NMP-TH), and arbitrary initial velocity $(\mathbf{v}_{0x}, \mathbf{v}_{0y})$ for parabolic throw (NMP-TP).




\noindent\textbf{(4-5) NMP-RD and NMP-RU: Ramp Sliding with Friction.}
For a ramp inclined by angle $\theta$ relative to the horizontal,  
$\hat{\mathbf{s}}=(t_x,t_y)$ is the unit tangent vector along the ramp’s downhill direction in the image plane, and $\hat{\mathbf{n}}$ the in-plane normal.  
Decomposing the gravitational force with 
kinetic friction $\mathbf{F}_f = -\mu_k {m} g \cos\theta\, \hat{\mathbf{s}}$, the net tangential force is 
\begin{equation}
\mathbf{F}_s = \pm\, ({m} g \sin\theta - \mu_k {m} g \cos\theta) \quad,
\end{equation}
where the positive/negative sign corresponds to sliding down/up.  
The constant tangential acceleration is
\begin{equation}
\mathbf{a}_s = \pm\, g(\sin\theta - \mu_k \cos\theta)  \quad,
\end{equation}
and its projection onto the image plane is $\mathbf{a} = (\mathbf{a}_s t_x, \mathbf{a}_s t_y)$.

%% file: sec/3_method_new.tex
\section{Method: \method}


\method post-trains video generators to follow physical laws. The main challenge lies in constructing reward losses that enforce \emph{physical constraints}. 
Given that the physics quantities are not directly measurable from generated videos, our core idea is to leverage \emph{measurable proxies} extracted from generated videos to construct these \emph{rewards}. 

\method{} has 
two components (Figure~\ref{fig:method}): (i) computing measurable proxies from model outputs (Section~\ref{sec:proxy}), and (ii) defining reward functions that quantify adherence to physics constraints; Section~\ref{sec:physics_constraints} details how these proxies are used to construct reward functions that enforce physical constraints. 
These reward signals are then used to fine-tune video generators, so that generated sequences not only appear realistic but also follow physical laws.

\subsection{Measurable Proxies}
\label{sec:proxy}
Let $G_\theta(\epsilon, c)$ denote a video generator parameterized by $\theta$, producing a video $V = G_\theta(\epsilon, c)$ given initial latent noise $\epsilon$ and condition $c$.
We define a measurable-proxy extractor $M$
which maps the generated video $V$ to a set of differentiable, physically meaningful quantities $M(V)$ (e.g., per-pixel displacement fields, object velocities, or visual features) that can serve as measurable proxies for physics quantities. 
In this work, we extract optical flow fields as proxies for velocity (Section~\ref{sec:velocity-proxy}) and visual appearance embeddings as proxies for mass-related properties (Section~\ref{sec:mass-proxy}). 
These proxies allow us to formulate physics-grounded 
constraints, such as constant-acceleration residuals and mass-conservation consistency,
that provide verifiable reward signals for fine-tuning the generator. 




\subsubsection{{Optical Flow as} Velocity Proxy
}
\label{sec:velocity-proxy}

In real-world videos we cannot directly observe $\mathbf{v}_t$.  
We instead employ an optical flow model $M_{OF}=\boldsymbol{\Psi}$ to estimate the per-frame displacement field using video frames $V_i$:
\begin{equation}
\boldsymbol{\phi}_t = \boldsymbol{\Psi}(V_t, V_{t+1})  \quad,
\end{equation}
where $\boldsymbol{\phi}_t\!=\!(\phi^x_t,\phi^y_t)$ denotes predicted optical flow (in pixels/frame).  
The image-plane velocity is approximated as:
\begin{equation}
\mathbf{v}_t \approx {\boldsymbol{\phi}_t}/{\Delta {t}}  \quad,
\label{eq:vel_proxy}
\end{equation}
with frame interval $\Delta {t} = 1/\mathrm{FPS}$.

\subsubsection{Visual Features as Mass Proxy
}
Newton’s Second Law states that acceleration is inversely proportional to mass under fixed external force (Eq.~\ref{eq:newton2}), implying that heavier objects exhibit smoother, slower changes in motion.
Though absolute mass is not directly observable in videos, appearance and texture cues often correlate with object identity, material, and thus effective mass.
We  define a \emph{mass proxy} based on high-level visual representations extracted from a pre-trained video encoder.
Let $\mathbf{z}_t = M_{\text{mass}}(V_t)$ denote the per-frame feature embedding obtained from the encoder, capturing consistent object-level appearance information over time.
This embedding space provides a differentiable, semantically aligned measure of visual mass consistency that can be compared across time or between simulated and generated videos.

\subsection{Physics Constraints and Rewards}
\label{sec:physics_constraints}
Let $\{C_j(m_j)\}_{j=1}^J$ denote a set of physical constraints or laws that should be satisfied by these extracted proxy quantities $m_j$.
We construct a physics penalty:
\begin{equation}
    \mathcal{L}_{\text{phys}}
    = \sum_{j=1}^J  \lambda_j \cdot \,  \ell\!\left(C_j[M_i(V)]\right)  \quad,
\label{eq:general_loss}
\end{equation}
where $\ell(\cdot)$ is a penalty or norm (e.g., squared error or hinge loss) and $\lambda_j$ are weighting coefficients.

\subsubsection{Discrete Constant-Acceleration Constraint}
\label{sec:mass-proxy}
For all primitives above, the image-plane accelerations $(\mathbf{a}_x, \mathbf{a}_y)$ remain constant throughout the motion.  
Discretizing the kinematic relation $\mathbf{v}_{t+1} = \mathbf{v}_t + \mathbf{a}\,\Delta {t}$ and $\mathbf{v}_{t} = \mathbf{v}_{t-1} + \mathbf{a}\,\Delta {t}$ yields
\begin{equation}
\mathbf{v}_{t+1} - 2\mathbf{v}_t + \mathbf{v}_{t-1} = \mathbf{0}  \quad.
\label{eq:constantacc}
\end{equation}
This discrete second-derivative constraint enforces the constant acceleration implied by Newton’s Second Law.

Substituting the proxy in Equation~\ref{eq:vel_proxy} into Equation~\ref{eq:constantacc}, we obtain the unified residual for all NMPs (see Section~\ref{sec:NMPs}):
\begin{equation}
\boldsymbol{\phi}_{t+1} - 2\,\boldsymbol{\phi}_t + \boldsymbol{\phi}_{t-1} \approx \mathbf{0}  \quad.
\end{equation}


\begin{proposition}[Newtonian Kinematic Constraint]
\label{prop:kinematic}
For an object governed by time-invariant external forces, the discrete second-order derivative of its optical-flow field predicted by $\boldsymbol{\Psi}$ vanishes:
\begin{equation}
\mathcal{R}_{\text{kinematic}} = \left\|
\boldsymbol{\phi}_{t+1} - 2\,\boldsymbol{\phi}_t + \boldsymbol{\phi}_{t-1}
\right\|_2^2 \approx \mathbf{0}  \quad.
\label{eq:phi_constraint}
\end{equation}
This is the optical-flow realization of Newton’s Second Law in the video domain, enforcing constant acceleration across all five Newtonian Motion Primitives.
\end{proposition}

Equation~\eqref{eq:phi_constraint} thus defines our \emph{constant-acceleration residual} $\mathcal{R}_{\text{kinematic}}$, capturing the essence of Newtonian dynamics in a differentiable, video-aligned form.

\subsubsection{Mass Conservation via Visual Features}


Given reference embeddings $\mathbf{z}^{\text{sim}}_t$ from physically simulated videos and $\mathbf{z}^{\text{gen}}_t$ from  generated videos, 
feature-level similarity constraint encourages mass preservation, 
so that the generator maintains consistent object appearance--hence consistent inferred mass--throughout the sequence and across domains. The mass conservation residual
\begin{equation}
\mathcal{R}_{\text{mass}} = 
\frac{1}{T} \sum_{t=0}^{T-1} 
\left\| 
\mathbf{z}^{\text{gen}}_t - \mathbf{z}^{\text{sim}}_t
\right\|_2^2 \quad,
\label{eq:massloss}
\end{equation}
 penalizes deviations between generated and simulated visual features.
Minimizing $\mathcal{R}_{\text{mass}}$ encourages the generator to produce motions and appearances that obey mass-dependent dynamics implied by Newton’s laws, complementing the kinematic constraint in Equation~\ref{eq:phi_constraint}.

\subsubsection{Post-Training Objective}

Our final objective combines a Newtonian kinematic constraint (Prop.~\ref{prop:kinematic}) and a mass-matching term (Eq.~\ref{eq:massloss}):
\begin{equation}
\mathcal{L}_{\text{phys}} = 
\lambda_{\text{kinematic}}\,\mathcal{R}_{\text{kinematic}}
+
\lambda_{\text{mass}}\,\mathcal{R}_{\text{mass}}  \quad.
\label{eq:final_loss}
\end{equation}
As the background is static and the camera is fixed in the construction of our dataset, the optical flow field $\boldsymbol{\phi}$ directly captures object motion, letting us compute the loss over the entire frame.  
This objective enforces Newtonian consistency without requiring explicit acceleration/depth supervision, unifying all five motion primitives (Section~\ref{sec:NMPs}) under a single principle:  under constant external forces, image-plane accelerations remain constant.

%% file: sec/4_dataset.tex
\section{NewtonBench-60K}

\label{sec:newtonbench}
We introduce \textbf{NewtonBench-60K}, a controlled, large-scale dataset and benchmark designed to isolate and evaluate five \emph{Newtonian Motion Primitives (NMPs)}: 
\emph{free fall}, \emph{horizontal throw}, \emph{parabolic throw}, \emph{sliding down a ramp with friction}, and \emph{sliding up a ramp with an uphill initial velocity}, each visualized in Figure~\ref{fig:dataset}. The corpus comprises \textbf{50K} simulated training videos (\textbf{10K} per NMP) and a \textbf{10K} held-out benchmark 
with \textbf{2K} videos per NMP, evenly split into \emph{In-Distribution} (ID) and \emph{Out-Of-Distribution} (OOD) subsets.
In comparison, PISA \cite{li2025pisa} focuses solely on free-fall and does not capture a broader range of Newtonian dynamics.

\subsection{Simulation Pipeline and Canonical Setups}
\label{subsec:sim-pipeline}
\noindent\textbf{Physics, Rendering, and Outputs.}
We build upon \textit{Kubric} for scene orchestration, \textit{PyBullet} for rigid-body dynamics, and \textit{Blender} for rendering. Gravity is $\mathbf{g}=(0,0,-9.81)$; videos are rendered at \textbf{512$\times$512} resolution, 
$32$ frames at $16$ fps with HDRI lighting. We adopt a fixed side-view camera for unambiguous 2D motion analysis; 
per-clip outputs include RGB frames, instance masks, depth maps, and metadata (camera intrinsics/extrinsics and object attributes).

\noindent\textbf{Assets and Splits.}
Objects are sampled from the GSO dataset~\cite{downs2022google} and backgrounds from HDRI \cite{greff2022kubric}. We create disjoint train/test pools for both objects and backgrounds, sampling strictly within the selected split.

\begin{table}[!ht]
\footnotesize
\centering
    \vspace{-6pt}
\caption{Natural motion primitive parameterization.}
    \vspace{-6pt}
\label{tab:nmp_params}
\resizebox{\linewidth}{!}{
\begin{tabular}{@{}p{0.24\columnwidth}p{0.76\columnwidth}@{}}
\toprule
\textbf{NMP} & \textbf{Description} \\
\midrule
Free fall & Spawn heights ${\sim}[0.5,\,1.5]$\,m; zero initial velocity. \\
\addlinespace[0.5ex]
Horizontal throw & Horizontal speed $v_0{\in}[2,\,6]$\,m/s; pitch $0^\circ$. \\
\addlinespace[0.5ex]
Parabolic throw & Speed $v_0\in[2,\,6]$\,m/s; launch angle $\theta\in[15^\circ,\,75^\circ]$. \\
\addlinespace[0.5ex]
Ramp slide down & Ramp angle $\theta\in[15^\circ,\,45^\circ]$, kinetic friction coefficients $(\mu_{\text{ramp}},\mu_{\text{obj}})\approx 0.06$; objects positioned near crest and released from rest. \\
\addlinespace[0.5ex]
Ramp slide up & Same ramp construction; initial uphill $v_0\in[3,\,4]$\,m/s imparted along the tangent. \\
\bottomrule
\end{tabular}
}
    \vspace{-6pt}
\end{table}

\noindent\textbf{NMP Parameterization.} 
 We generate each primitive in Table~\ref{tab:nmp_params}, by sampling a small set of physically meaningful parameters; 
ramps are built from rectangular colliders (stable physics) and visually aligned slabs (rendering), with top plane and downhill direction computed from the ramp mesh for consistent tangential placement and motion direction.






\noindent\textbf{Mask Extraction for Evaluation.}
For ground truth, we use renderer instance masks to obtain per-frame object regions. 
For generated videos, we extract object masks with {SAM2}~\cite{Ravi_2025_Sam2} guided by conditioning frames and object prompts;
centroids $\mathbf{c}^{\text{gen}}$ are then computed from these masks for metrics evaluation (Sec.~\ref{subsec:metrics}). 

\begin{table*}[!t]
\vspace{-15pt}
\footnotesize
\caption{Comparison of different post-training strategies on the OpenSora (SFT) baseline. Percentages indicate relative change vs. baseline ({\textcolor{ForestGreen}{green}} = improvement, {\textcolor{BrickRed}{red}} = regression). Visual metrics (L2, CD, IoU) capture pixel alignment and shape agreement; physics metrics ($\mathrm{RMSE}_{\mathbf{v}}$, $\mathrm{RMSE}_{\mathbf{a}}$) capture physical plausibility in motion.}
\vspace{-6pt}
\centering
\setlength{\tabcolsep}{4pt}
\resizebox{\linewidth}{!}{
\begin{tabular}{lllllll}
\toprule
& \multicolumn{3}{c}{\textbf{Visual metrics}} & \multicolumn{2}{c}{\textbf{Physics metrics}} &  \\
\textbf{Method} & \textbf{L2} ($\downarrow$) & \textbf{CD} ($\downarrow$) & \textbf{IoU} ($\uparrow$) & $\mathrm{\mathbf{RMSE}}_{\mathbf{v}}$
 ($\downarrow$) & $\mathrm{\mathbf{RMSE}}_{\mathbf{a}}$ ($\downarrow$) & \textbf{Avg. Change} \\
\midrule
OpenSora (SFT) & 0.1098 & 0.3159 & 0.1103 & 0.2792 & 3.3244 & -- \\
\midrule

PISA~\cite{li2025pisa} ORO Optical Flow
& 0.1042 {\small(\textcolor{ForestGreen}{+5.10\%})}
& 0.2963 {\small(\textcolor{ForestGreen}{+6.18\%})}
& 0.1179 {\small(\textcolor{ForestGreen}{+6.88\%})}
& 0.2799 {\small(\textcolor{BrickRed}{-0.25\%})}
& 2.7217 {\small(\textcolor{ForestGreen}{+18.12\%})}
& \textcolor{ForestGreen}{+7.61\%} \\

PISA~\cite{li2025pisa} ORO Depth Map
& 0.1079 {\small(\textcolor{ForestGreen}{+1.73\%})}
& 0.3114 {\small(\textcolor{ForestGreen}{+1.43\%})}
& 0.1146 {\small(\textcolor{ForestGreen}{+3.90\%})}
& 0.2875 {\small(\textcolor{BrickRed}{-2.97\%})}
& 3.4652 {\small(\textcolor{BrickRed}{-4.23\%})}
& \textcolor{ForestGreen}{+0.37\%} \\

PISA~\cite{li2025pisa} ORO Segmentation
& 0.1099 {\small(\textcolor{BrickRed}{-0.09\%})}
& 0.3177 {\small(\textcolor{BrickRed}{-0.57\%})}
& 0.1138 {\small(\textcolor{ForestGreen}{+3.17\%})}
& 0.2796 {\small(\textcolor{BrickRed}{-0.14\%})}
& 3.2943 {\small(\textcolor{ForestGreen}{+0.91\%})}
& \textcolor{ForestGreen}{+0.65\%} \\

\rowcolor{gray!30}\method & \textbf{0.0962} {\small(\textcolor{ForestGreen}{+12.39\%})}
        & \textbf{0.2930} {\small(\textcolor{ForestGreen}{+7.25\%})}
        & \textbf{0.1266} {\small(\textcolor{ForestGreen}{+14.78\%})}
        & \textbf{0.2628} {\small(\textcolor{ForestGreen}{+5.87\%})}
        & 3.0432 {\small(\textcolor{ForestGreen}{+8.46\%})}
        & \textbf{\textcolor{ForestGreen}{+9.75\%}} \\
\method (w/o residual)
        & 0.1109 {\small(\textcolor{BrickRed}{-1.00\%})}
        & 0.3199 {\small(\textcolor{BrickRed}{-1.27\%})}
        & 0.1145 {\small(\textcolor{ForestGreen}{+3.81\%})}
        & 0.2793 {\small(\textcolor{BrickRed}{-0.04\%})}
        & 3.3321 {\small(\textcolor{BrickRed}{-0.23\%})}
        & \textcolor{ForestGreen}{+0.25\%} \\
\method (w/o mass)
        & 0.1055 {\small(\textcolor{ForestGreen}{+3.92\%})}
        & 0.2993 {\small(\textcolor{ForestGreen}{+5.26\%})}
        & 0.1165 {\small(\textcolor{ForestGreen}{+5.62\%})}
        & 0.2737 {\small(\textcolor{ForestGreen}{+1.97\%})}
        & \textbf{2.5348} {\small(\textcolor{ForestGreen}{+23.75\%})}
        & \textcolor{ForestGreen}{+8.10\%} \\
\bottomrule
\end{tabular}
}
\label{tab:gravity_rewards}
\vspace{-8pt}
\end{table*}

\subsection{Benchmark Protocol: ID \& OOD}
\label{subsec:benchmark}
For each NMP, we synthesize \textbf{1K ID} and \textbf{1K OOD} videos. 
ID parameter ranges mirror training (e.g., fall height $[0.5,1.5]$\,m; throw speed $[2,6]$\,m/s; ramp angle $[15^\circ,45^\circ]$). 
OOD ranges deliberately hold out disjoint bands to stress generalization: 
\emph{higher} horizontal throws (e.g., $[1.7,2.0]$\,m/s),
\emph{higher} parabolic throws (e.g., $[1.7,2.0]$\,m/s), \emph{extreme} parabolic angles (e.g., $(75^\circ,90^\circ]$),
and \emph{steeper/shallower} ramps (e.g., $(45^\circ,60^\circ]$ with \emph{faster} sliding up $[4.0,5.0]$\,m/s). 
We optionally perturb friction by $\pm 25\%$ in OOD to decouple appearance from dynamics.

\subsection{Evaluation Metrics}
\label{subsec:metrics}

All metrics are computed per object (frame-aligned), then averaged across objects and videos. 
For frame interval $\Delta t$, $\mathbf{c}^{\text{gen}}_t, \mathbf{c}^{\text{gt}}_t \in \mathbb{R}^2$ are generated and ground-truth centroids at frame $t$.
Evaluation includes physics-based metrics (velocity and acceleration RMSE) and standard visual metrics (L2, CD, and IoU) on both in and out-of distribution splits.

\noindent\textbf{Velocity RMSE.}
We define image-plane velocities by the first discrete derivative $\mathbf{v}_t \;=\; \frac{\mathbf{c}_{t+1}-\mathbf{c}_t}{\Delta t}\,.$  Our velocity error measures Newtonian consistency of first-order kinematics:
$
\mathrm{\textbf{RMSE}}_{\mathbf{v}}
\;=\;
\sqrt{\frac{1}{T-1}\sum_{t=1}^{T-1}\left\|\mathbf{v}^{\text{gen}}_t-\mathbf{v}^{\text{gt}}_t\right\|_2^2}.
$

\noindent\textbf{Acceleration RMSE.}
Likewise, we define image-plane accelerations by the discrete second-order derivative,
\[
\mathbf{a}_t \;=\; \frac{\mathbf{v}_{t+1}-\mathbf{v}_t}{\Delta t}
\;=\; \frac{\mathbf{c}_{t+2}-2\mathbf{c}_{t+1}+\mathbf{c}_t}{\Delta t^2}\,  \quad,
\]
we report
$
\mathrm{\textbf{RMSE}}_{\mathbf{a}}
\;=\;
\sqrt{\frac{1}{T-2}\sum_{t=1}^{T-2}\left\|\mathbf{a}^{\text{gen}}_t-\mathbf{a}^{\text{gt}}_t\right\|_2^2}.
$
These two physics metrics directly evaluate if generated motions obey constant-acceleration behavior in both axes.

\noindent\textbf{Standard Visual Metrics (following PISA).}
Spatial fidelity metrics 
\emph{Trajectory Position Error (centroid L2)} 
$\mathrm{L2\_traj}=\frac{1}{T}\sum_{t=1}^{T}\|\mathbf{c}^{\text{gen}}_t-\mathbf{c}^{\text{gt}}_t\|_2$,
\emph{Chamfer Distance} (CD) between binary masks (per frame), and \emph{Intersection over Union} (IoU) (per-frame overlap), 
capture pixel-space alignment and shape agreement.

%% file: sec/5_experiment.tex
\section{Experiments}
\label{sec:experiments}
\subsection{Experimental Settings}
We fine-tune our base text-to-video diffusion model Open-Sora v1.2~\cite{zheng2024open} on our \texttt{NewtonBench-60K} dataset comprising 50K simulated videos across five NMPs, and fine-tune it to accept both text and the first 4 frames of a video as conditions. Both Supervised Fine-Tuning (SFT) and post-training operate on 32-frame clips at 16\,fps, consistent with the dataset specification. Training is performed on 8$\times$NVIDIA H100 (80GB) GPUs with a batch size of 1 and gradient accumulation of 32. The learning rate is set to $1\times10^{-4}$ for SFT and $1\times10^{-5}$ for post-training. We employ the RAFT~\cite{teed2020raft} optical-flow model to compute motion fields and the V-JEPA~2~\cite{assran2025v} encoder to extract visual features for mass alignment. Evaluation follows the \texttt{NewtonBench-60K} protocol.

\subsection{Experimental Results}

\textbf{Comparison with State of The Art.}  
We compare our \method framework with three post-training strategies adapted from PISA~\cite{li2025pisa}: \emph{Optical Flow Reward}, \emph{Depth Reward}, and \emph{Segmentation Reward}. All methods are fine-tuned from the same OpenSora (SFT) baseline under identical settings on our \texttt{NewtonBench-60K} dataset. In PISA, each reward measures the similarity between the generated video and its simulated ground truth: RAFT~\cite{teed2020raft} computes optical flow fields and minimizes their discrepancy, Depth-Anything-V2~\cite{yang2024depth} aligns predicted and true depth maps, and SAM2~\cite{Ravi_2025_Sam2} provides object masks for IoU-based supervision.  
Table~\ref{tab:gravity_rewards} reports the results. 

We observe that visual feature-based rewards improve appearance metrics, \ie, Depth and Optical flow rewards slightly enhance L2, CD, and IoU; our Mass reward improves IoU. However, they do not guarantee physically consistent motion--reflected in higher velocity and/or acceleration errors ($\mathrm{RMSE}_{\mathbf{v}}$, $\mathrm{RMSE}_{\mathbf{a}}$). 
\begin{tcolorbox}[colback=white,colframe=black!70!white,boxrule=0.5pt,left=5pt,right=5pt,top=2pt,bottom=2pt]
\textbf{Finding 1.} \textit{Visual feature alignment improves perceptual and spatial fidelity but fails to enforce adherence to physical laws of motion.}
\end{tcolorbox}
As shown in Table~\ref{tab:gravity_rewards}, PISA ORO--Depth shows small spatial gains ($+1$--$4\%$) but degrades temporal consistency ($-3\%$ in $\mathrm{RMSE}_{\mathbf{v}}$, $-4\%$ in $\mathrm{RMSE}_{\mathbf{a}}$). Post-training only with our mass conservation reward shows a similar trend (row \method w/o residual). Segmentation and optical flow rewards slightly worsen velocity measures, suggesting that frame-level feature alignment alone cannot capture Newtonian dynamics.
In contrast, \method achieves consistent improvements across all five metrics (average $+9.75\%$), demonstrating that enforcing self-consistent kinematic and mass constraints provides a stronger inductive bias for physically grounded and temporally coherent video generation.

\begin{figure}[t!]
    \centering
    \includegraphics[width=\linewidth]{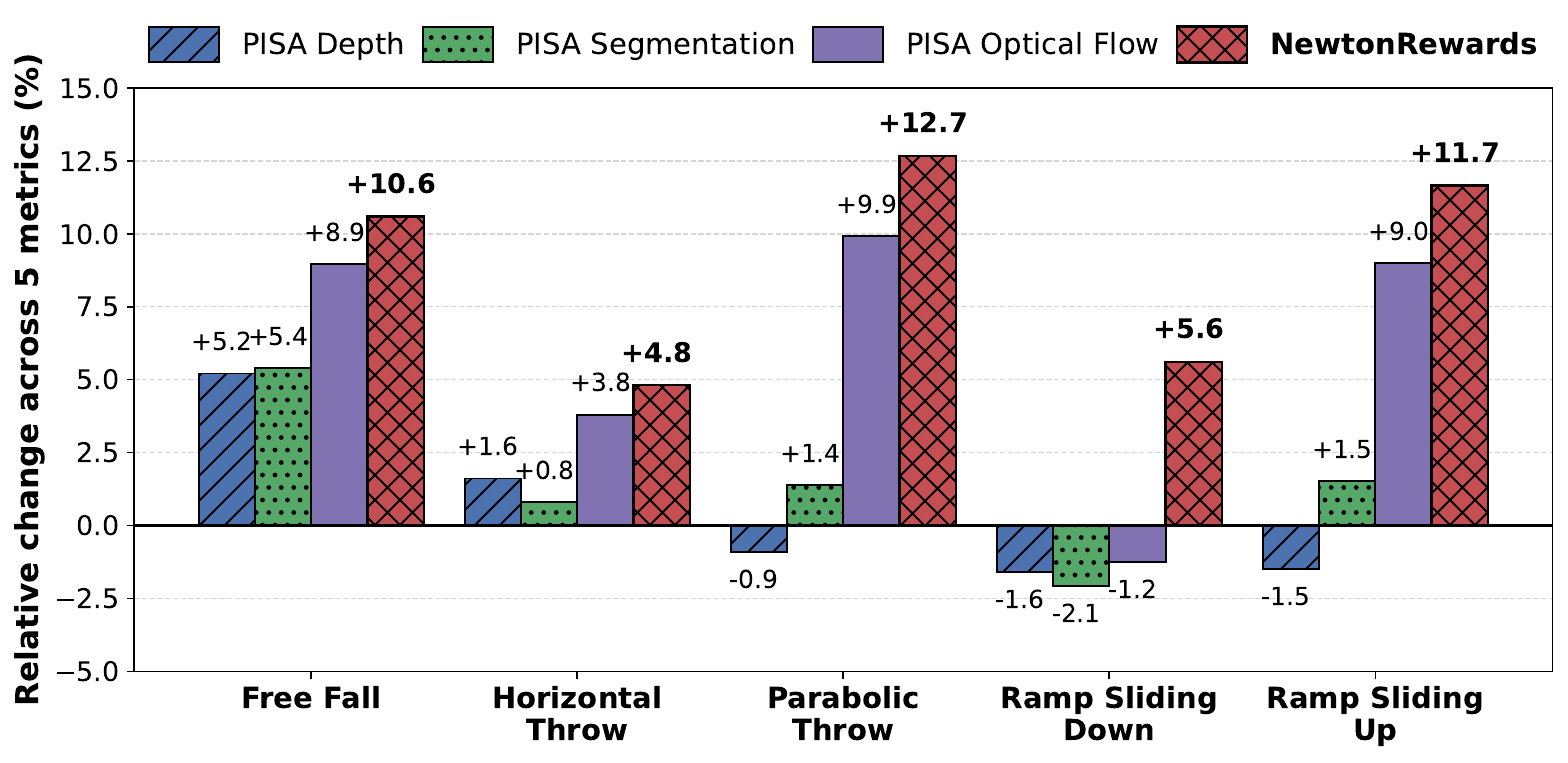}
    \vspace{-5mm}
    \caption{\textbf{Relative performance change across Newtonian Motion Primitives.}
Percentage improvements over the SFT baseline across all five NMPs. Depth and Segmentation provide modest gains on simple motions but degrade on ramp dynamics, while Optical Flow shows highly variable and unstable behavior. In contrast, \method delivers consistent positive improvements across all primitives, demonstrating robust generalization to diverse Newtonian dynamics.}
\label{fig:nmp_changes}
\vspace{-14pt}
\end{figure}

\noindent\textbf{Evaluation Across Newtonian Motion Primitives.} Fig.~\ref{fig:nmp_changes} shows the relative performance gain of each post-training strategy over the OpenSora (SFT) baseline across all five NMPs. PISA rewards show inconsistent and task-dependent behavior: Depth is mildly better on free fall and horizontal throw, but worse on motions such as parabolic and ramp dynamics. Same with Segmentation--small gains on simple trajectories but clear regression on ramp-down and marginal improvement elsewhere. Optical Flow is highly unstable, with large swings across primitives, including strong regression on ramp-down despite large gains on free fall and parabolic throw. In contrast, \method provides \emph{uniformly positive} improvements across all five motion primitives. Its largest gains are on the most challenging motions--up to \textbf{+12.7\%} on parabolic throws and \textbf{+11.7\%} on ramp-up--and still outperforms all baselines on simple trajectories. 
Enforcing Newtonian kinematics and mass consistency yields robust, cross-regime improvements that generalize reliably across diverse physical scenarios.

\begin{table*}[!t]
\vspace{-16pt}
\footnotesize
\caption{Out-of-distribution (OOD) evaluation on 5K OOD benchmark of \texttt{NewtonBench-60K}. 
\method improves consistently across all metrics, demonstrating stronger generalization than the OpenSora (SFT) baseline.}
\vspace{-8pt}
\centering
\setlength{\tabcolsep}{4pt}
\resizebox{\linewidth}{!}{
\begin{tabular}{lllllll}
\toprule
\textbf{Method} & \textbf{L2} ($\downarrow$) & \textbf{CD} ($\downarrow$) & \textbf{IoU} ($\uparrow$) & $\mathrm{\mathbf{RMSE}}_{\mathbf{v}}$
 ($\downarrow$) & $\mathrm{\mathbf{RMSE}}_{\mathbf{a}}$ ($\downarrow$) & \textbf{Avg. Change} \\
\midrule
OpenSora (SFT) -- ID & 0.1098 & 0.3159 & 0.1103 & 0.2792 & 3.3244 & -- \\
\rowcolor{gray!30}\method -- ID & \textbf{0.0962} {\small(\textcolor{ForestGreen}{+12.39\%})}
        & \textbf{0.2930} {\small(\textcolor{ForestGreen}{+7.25\%})}
        & \textbf{0.1266} {\small(\textcolor{ForestGreen}{+14.78\%})}
        & \textbf{0.2628} {\small(\textcolor{ForestGreen}{+5.87\%})}
        & \textbf{3.0432} {\small(\textcolor{ForestGreen}{+8.46\%})}
        & \textbf{\textcolor{ForestGreen}{+9.75\%}} \\
\midrule
OpenSora (SFT) -- OOD & 0.1297 & 0.4082 & 0.0998 & 0.4230 & 6.1451 & -- \\

\rowcolor{gray!30}\method -- OOD 
& \textbf{0.1206} {\small(\textcolor{ForestGreen}{+7.02\%})}
& \textbf{0.3780} {\small(\textcolor{ForestGreen}{+7.40\%})}
& \textbf{0.1025} {\small(\textcolor{ForestGreen}{+2.71\%})}
& \textbf{0.3816} {\small(\textcolor{ForestGreen}{+9.79\%})}
& \textbf{5.1561} {\small(\textcolor{ForestGreen}{+16.09\%})}
& \textbf{\textcolor{ForestGreen}{+8.60\%}} \\
\bottomrule
\end{tabular}
}
\label{tab:ood}
\vspace{-14pt}
\end{table*}

\begin{figure}[t!]
    \centering
    \includegraphics[width=\linewidth]{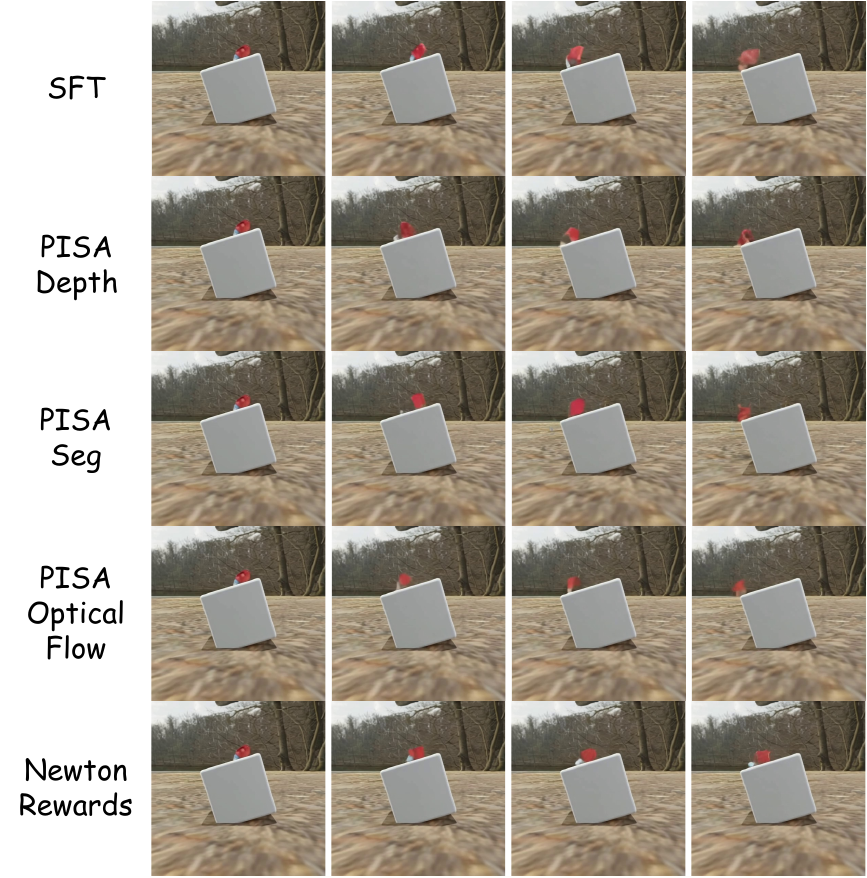}
    \vspace{-7mm}
    \caption{
\textbf{Qualitative comparison of post-training strategies on the \texttt{NewtonBench-60K} ramp-slide down scenario.} 
Clear differences emerge when inspecting the temporal evolution across frames (left→right).
For SFT and all PISA variants (Depth, Seg, Optical Flow), the cube exhibits inconsistent deceleration and unstable surface contact-evident in Frames 2–4, where the cube tilts unnaturally, slips erratically, or momentarily “floats’’ above the ramp.
PISA Optical Flow especially shows noticeable jitter and non-smooth frame-to-frame motion.
In contrast, \method maintains stable grounding and smooth, constant-acceleration motion across all frames. 
}
\label{fig:qualitative}
\vspace{-15pt}
\end{figure}

\begin{figure}[t!]
    \centering
    \includegraphics[width=\linewidth]{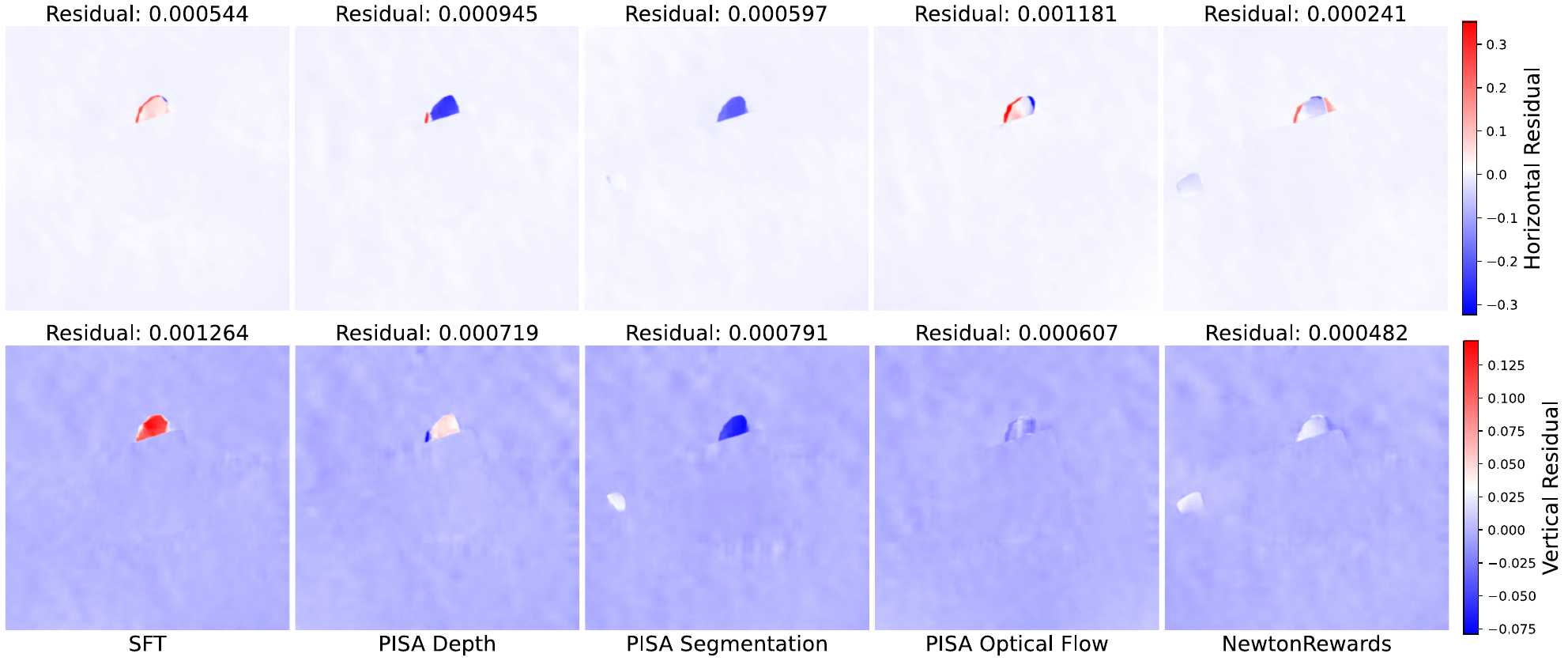}
    \vspace{-15pt}
    \caption{Mean horizontal and vertical residuals $\boldsymbol{\phi}_{t+1} - 2\,\boldsymbol{\phi}_t + \boldsymbol{\phi}_{t-1}$. Lower magnitude indicates closer adherence to constant-acceleration dynamics. \method produces the smallest residuals, while SFT and PISA variants show larger deviations.}   
\label{fig:residual}
\vspace{-14pt}
\end{figure}

\noindent\textbf{Qualitative Comparison.}
Representative results across different post-training strategies are in Fig.~\ref{fig:qualitative}. 
While PISA-based rewards grounded in visual similarity (Depth, Segmentation, Optical Flow) sometimes improve local appearance, they fail to enforce physically coherent motion: objects either drift unnaturally or exhibit inconsistent acceleration when interacting with the ramp. For example, PISA Depth (Row 2) shows the cube briefly losing contact in Frame 3, and PISA OF (Row 4) produces a sudden orientation snap between Frames 3-4. In contrast, our \method yields visually realistic and physically consistent trajectories--objects maintain stable contact, decelerate smoothly under friction, and adhere to Newtonian expectations. Visual observations and quantitative trends in Table~\ref{tab:gravity_rewards} and Fig.~\ref{fig:residual}, confirm that physics-grounded verifiable rewards promote perceptual fidelity and dynamic realism.

\noindent\textbf{OOD Evaluation.} Table~\ref{tab:ood} evaluates generalization under distribution shift, where test videos exhibit higher drop heights, faster throws, steeper ramps, and perturbed friction compared to training. The OpenSora (SFT) baseline degrades substantially in this setting (e.g., L2 increases to 0.1297 and acceleration error nearly doubles to 6.15), reflecting limited robustness to unseen physical configurations. In contrast, \method consistently improves across all five metrics--achieving a +7.01\% reduction in L2, +7.38\% improvement in CD, and +9.79\% reduction in acceleration error; these, despite never observing OOD dynamics during post-training. These clearly show that enforcing Newtonian kinematic and mass constraints yields models that not only fit training physics more faithfully but also extrapolate more reliably to unseen physical regimes.

\noindent\textbf{Constant-Acceleration Residual Analysis.} To directly assess whether generated motions obey Newtonian kinematics, we compute the mean discrete second-order residual $\boldsymbol{\phi}_{t+1} - 2\,\boldsymbol{\phi}_t + \boldsymbol{\phi}_{t-1}$, averaged over all 32 frames of the sliding-down-ramp scenario for each method, as in Figure~\ref{fig:qualitative}. This residual is zero for ideal constant-acceleration motion and therefore serves as a sensitive diagnostic of dynamical consistency. Figure~\ref{fig:residual} shows the horizontal (top) and vertical (bottom) residual fields. The SFT baseline and all PISA variants produce strong \textcolor{red}{red}/\textcolor{blue}{blue} activations, indicating noticeable violations of the constant-acceleration constraint. Even methods that use ground-truth visual signals (PISA Depth, Segmentation, and Optical Flow) retain substantial structured residuals, revealing that pixel-level alignment does not translate into correct governing dynamics. In contrast, \method produces markedly smoother residual maps with minimal magnitude, achieving the lowest absolute residuals across both axes. These reductions demonstrate that enforcing Newtonian kinematic structure yields trajectories that more closely adhere to true constant-acceleration behavior, beyond what can be captured through appearance- or flow-based supervision alone.
 
\begin{figure}[t!]
    \centering
    \includegraphics[width=\linewidth]{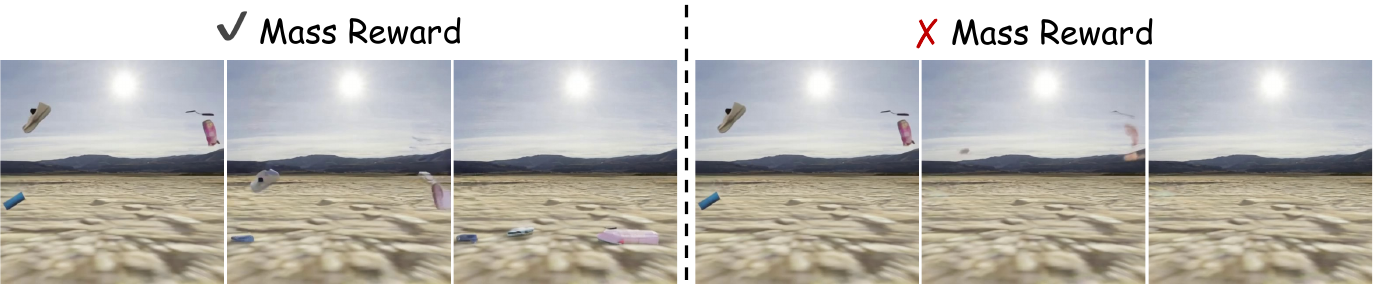}
    \vspace{-18pt}
    \caption{Videos generated with the mass reward (left) maintain consistent object persistence, while removing it (right) leads to degenerate behavior where objects vanish; an instance of reward hacking when optimizing only the kinematic residual.}
    \label{fig:hacking}
\vspace{-14pt}
\end{figure}

\subsection{Ablation Study}
\noindent\textbf{Newtonian Kinematic Residual Constraint.} Ablating the discrete residual term largely removes the temporal regularization effect, leading to only marginal overall improvement over SFT (+0.25\%). Without enforcing constant acceleration, the model produces visually coherent but physically inconsistent motion, i.e., slightly better spatial fidelity (IoU +3.8\%) yet degraded kinematic accuracy across L2, CD, velocity, and acceleration metrics. This highlights that the residual constraint is essential for stabilizing motion and aligning generated dynamics with Newtonian laws.

\noindent\textbf{Mass Conservation Reward.}
Although the overall gain slightly decreases (+8.1\% \!vs.\! +9.75\%) when removing mass conservation, as seen in Table~\ref{tab:gravity_rewards}, this configuration only employs the Newtonian kinematic residual without additional regularization. 
To understand its influence, we assess the role of mass conservation in stabilizing post-training.
\begin{tcolorbox}[colback=white,colframe=black!70!white,boxrule=0.5pt,left=5pt,right=5pt,top=2pt,bottom=2pt]
\textbf{Finding 2.} \textit{The mass conservation reward mitigates reward hacking that emerges when optimizing solely the kinematic residual, thereby avoiding trivial solutions}.
\end{tcolorbox}
Without this constraint, the model converges to a degenerate, trivial solution that minimizes the residual by driving all velocities to zero (\(\mathbf{v}_t = 0\)), effectively causing object disappearance. 
By anchoring visual and feature-level mass consistency, the mass reward prevents this collapse and ensures stable, physically meaningful motion as in Figure~\ref{fig:hacking}.

%% file: sec/6_conclusion.tex
\section{Conclusion}
\label{sec:conclusion}


We introduced \method, a general physics-grounded post-training framework that enforces Newtonian consistency in video generation by constructing verifiable rewards from measurable proxies. By leveraging optical flow and visual features as surrogates for velocity and mass, \method enforces Newtonian dynamics through kinematic and mass-conservation constraints. Our method enables video generators to obey constant-acceleration dynamics and maintain physical plausibility across diverse NMPs. Beyond Newtonian mechanics, our approach is inherently general. This unlocks a broader view of physics-grounded post-training: once a physical quantity can be estimated, generative models can be guided toward physically valid behavior through explicit, differentiable constraints.

%% file: sec/X_suppl.tex
\clearpage
\setcounter{page}{1}
\maketitlesupplementary
\noindent This supplementary document provides additional results and analyses supporting the main paper. 
Section~\ref{sec:realworld} presents a \emph{real-world evaluation} on 361 free-fall videos from the PISA~\cite{li2025pisa} benchmark, showing that our physics-grounded post-training--developed entirely in simulation--transfers to natural scenes and real gravitational motion. 
Section~\ref{sec:reward_hacking} provides \emph{additional quantitative evidence} for Finding~2, demonstrating mass conservation reward prevents reward hacking by avoiding degenerate zero-motion solutions. 
Section~\ref{sec:extended_qualitative} offers \emph{extended qualitative comparisons} across the remaining NMPs, with frame-level visualizations highlighting the improved consistency and stability of \method. 
We also include \emph{video-based qualitative results}: comparisons for all NMPs and real-world sequences at multiple frame rates (16\,fps, 8\,fps, 4\,fps), and the teaser video from the main paper (see attached videos).
\section{Real-World Experiment}
\label{sec:realworld}

\begin{figure}[h]
    \centering
    \includegraphics[width=\linewidth]{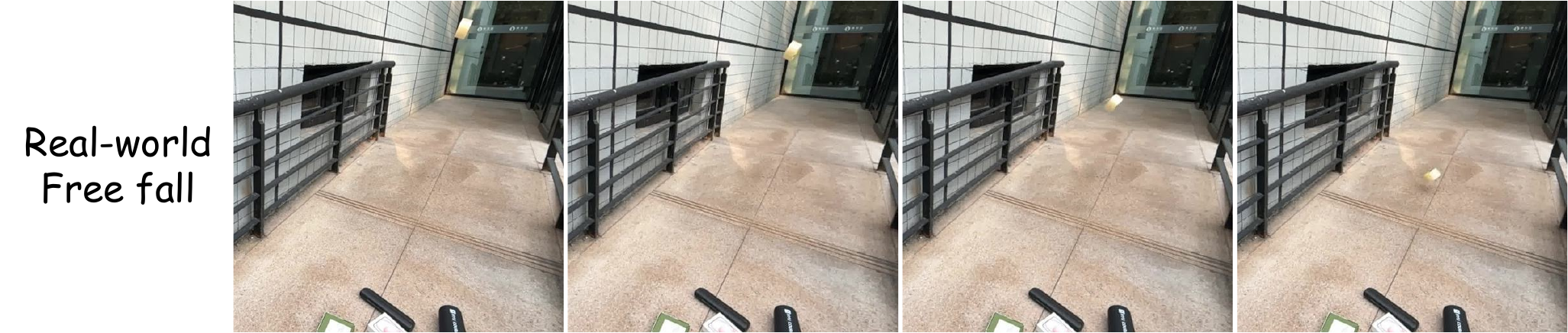}
    \caption{\textbf{Real-world free-fall evaluation.}  
We test whether models post-trained purely in simulation can generalize to real camera imagery and real gravitational motion. Shown here is a representative video from the PISA real-world dataset (361 free-fall videos).}
\label{fig:realworld}
\end{figure}

To test whether physics-grounded post-training transfers beyond controlled simulation, we evaluate our approach on 361 real-world free-fall videos provided in the PISA benchmark~\cite{li2025pisa}. These videos capture everyday objects dropped in diverse indoor and outdoor environments, providing natural variation in texture, lighting, background clutter, and real gravitational motion (see Figure~\ref{fig:realworld}). This setting allows us to assess whether a model trained purely on synthetic physics signals can generalize to real camera imagery and real-world dynamics. 

Following the evaluation protocol used in our \texttt{NewtonBench-60K}, we measure performance using five metrics: three visual metrics (L2, Chamfer Distance, IoU) and two physics metrics ($\mathrm{RMSE}_{\mathbf{v}}$, $\mathrm{RMSE}_{\mathbf{a}}$). We evaluate the OpenSora (SFT) baseline, all three PISA post-training variants (Depth, Segmentation, Optical Flow), and our \method under exactly the same conditions. This establishes a direct comparison of how well different reward formulations cope with real free-fall trajectories.

As shown in Table~\ref{tab:realworld}, \method yields the largest and maintains consistent gains across both visual and physics metrics, demonstrating strong transfer from synthetic supervision to natural free-fall motion.

\section{Reward Hacking Mitigation}
\label{sec:reward_hacking}
\begin{table}[!h]
\centering
\footnotesize
\caption{\textbf{Residuals and velocity magnitude.}
Lower residuals indicate smoother (closer-to-constant-acceleration) motion, while the velocity magnitude reveals whether the motion remains physically meaningful. Without mass conservation, the model reduces the residual primarily by collapsing motion magnitude.}
\resizebox{\linewidth}{!}{
\begin{tabular}{lccc}
\toprule
\textbf{Method} 
& \textbf{\shortstack{Horizontal\\Residual}} 
& \textbf{\shortstack{Vertical\\Residual}} 
& \textbf{\shortstack{Velocity\\Magnitude}} \\
\midrule
OpenSora (SFT)      & 0.000509 & 0.001142 & 0.101715 \\
\method             & \textbf{0.000194} & 0.000511 & 0.071377 \\
\method w/o mass    & 0.000986 & \textbf{0.000331} & \textcolor{BrickRed}{\textbf{0.033854}} \\
\bottomrule
\end{tabular}
}
\label{tab:reward_hacking}
\end{table}

In Section~\ref{sec:experiments} of the main paper, we identified \emph{reward hacking} as a failure mode that arises when optimizing only the kinematic residual.  
Without the mass-conservation reward, the generator can trivially reduce
\(
\left\|
\boldsymbol{\phi}_{t+1} - 2\,\boldsymbol{\phi}_t + \boldsymbol{\phi}_{t-1}
\right\|_2^2
\)
by driving all velocity fields $\boldsymbol{\phi}_t$ toward zero--producing videos in which the object barely moves or even disappears.  
Here, we provide additional quantitative evidence of this finding.

As shown in Table~\ref{tab:reward_hacking}, removing mass conservation (\method w/o mass) decreases the vertical residual compared to SFT, but does so by collapsing the average velocity magnitude from $0.1017$ to $0.0339$--a reduction of more than $66\%$.  
This confirms that the residual-only model optimizes the objective by freezing motion rather than by producing more accurate Newtonian trajectories.  
In contrast, the full \method not only yields substantially lower residuals in both directions, but also maintains non-trivial velocity, indicating that it improves dynamical consistency without sacrificing meaningful motion.  
These results quantitatively support our claim that the mass-conservation reward is crucial for preventing reward hacking and stabilizing physics-grounded post-training.

\begin{table*}[h]
\footnotesize
\caption{SFT and post-training strategies on real-world experiment. ({\textcolor{ForestGreen}{green}} = improvement, {\textcolor{BrickRed}{red}} = regression). Visual metrics (L2, CD, IoU) capture pixel alignment and shape agreement; physics metrics ($\mathrm{RMSE}_{\mathbf{v}}$, $\mathrm{RMSE}_{\mathbf{a}}$) capture physical plausibility in motion.}
\vspace{-3mm}
\centering
\setlength{\tabcolsep}{4pt}
\resizebox{0.8\linewidth}{!}{
\begin{tabular}{lllllll}
\toprule
& \multicolumn{3}{c}{\textbf{Visual metrics}} & \multicolumn{2}{c}{\textbf{Physics metrics}} &  \\
\textbf{Method} & \textbf{L2} ($\downarrow$) & \textbf{CD} ($\downarrow$) & \textbf{IoU} ($\uparrow$) & $\mathrm{\mathbf{RMSE}}_{\mathbf{v}}$ ($\downarrow$) & $\mathrm{\mathbf{RMSE}}_{\mathbf{a}}$ ($\downarrow$) & \textbf{Avg. Change} \\
\midrule
OpenSora (SFT) & 0.1716 & 0.4386 & 0.0198 & 2.4485 & 18.4169 & -- \\
\midrule

PISA~\cite{li2025pisa} ORO Optical Flow
& 0.1699 {\small(\textcolor{ForestGreen}{+0.99\%})}
& 0.4336 {\small(\textcolor{ForestGreen}{+1.14\%})}
& 0.0182 {\small(\textcolor{BrickRed}{-8.08\%})}
& 2.4237 {\small(\textcolor{ForestGreen}{+1.01\%})}
& 18.3333 {\small(\textcolor{ForestGreen}{+0.45\%})}
& \textcolor{BrickRed}{-0.90\%} \\

PISA~\cite{li2025pisa} ORO Depth Map
& 0.1704 {\small(\textcolor{ForestGreen}{+0.70\%})}
& 0.4342 {\small(\textcolor{ForestGreen}{+1.00\%})}
& 0.0218 {\small(\textcolor{ForestGreen}{+10.10\%})}
& 2.4395 {\small(\textcolor{BrickRed}{-0.37\%})}
& 18.3474 {\small(\textcolor{BrickRed}{-0.38\%})}
& \textcolor{ForestGreen}{+2.21\%} \\

PISA~\cite{li2025pisa} ORO Segmentation
& 0.1712 {\small(\textcolor{ForestGreen}{+0.23\%})}
& 0.4372 {\small(\textcolor{ForestGreen}{+0.32\%})}
& 0.0210 {\small(\textcolor{ForestGreen}{+6.06\%})}
& 2.4273 {\small(\textcolor{ForestGreen}{+0.87\%})}
& 18.2344 {\small(\textcolor{ForestGreen}{+0.99\%})}
& \textcolor{ForestGreen}{+1.29\%} \\

\rowcolor{gray!30}\method 
& \textbf{0.1698} {\small(\textcolor{ForestGreen}{+1.05\%})}
& \textbf{0.4333} {\small(\textcolor{ForestGreen}{+1.21\%})}
& \textbf{0.0235} {\small(\textcolor{ForestGreen}{+18.69\%})}
& \textbf{2.3889} {\small(\textcolor{ForestGreen}{+2.43\%})}
& \textbf{18.1670} {\small(\textcolor{ForestGreen}{+1.36\%})}
& \textbf{\textcolor{ForestGreen}{+4.15\%}} \\
\bottomrule
\end{tabular}
}
\label{tab:realworld}
\vspace{-4mm}
\end{table*}

\section{Extended Qualitative Results}
\label{sec:extended_qualitative}
\begin{figure*}
\centering
\includegraphics[width=0.8\linewidth]{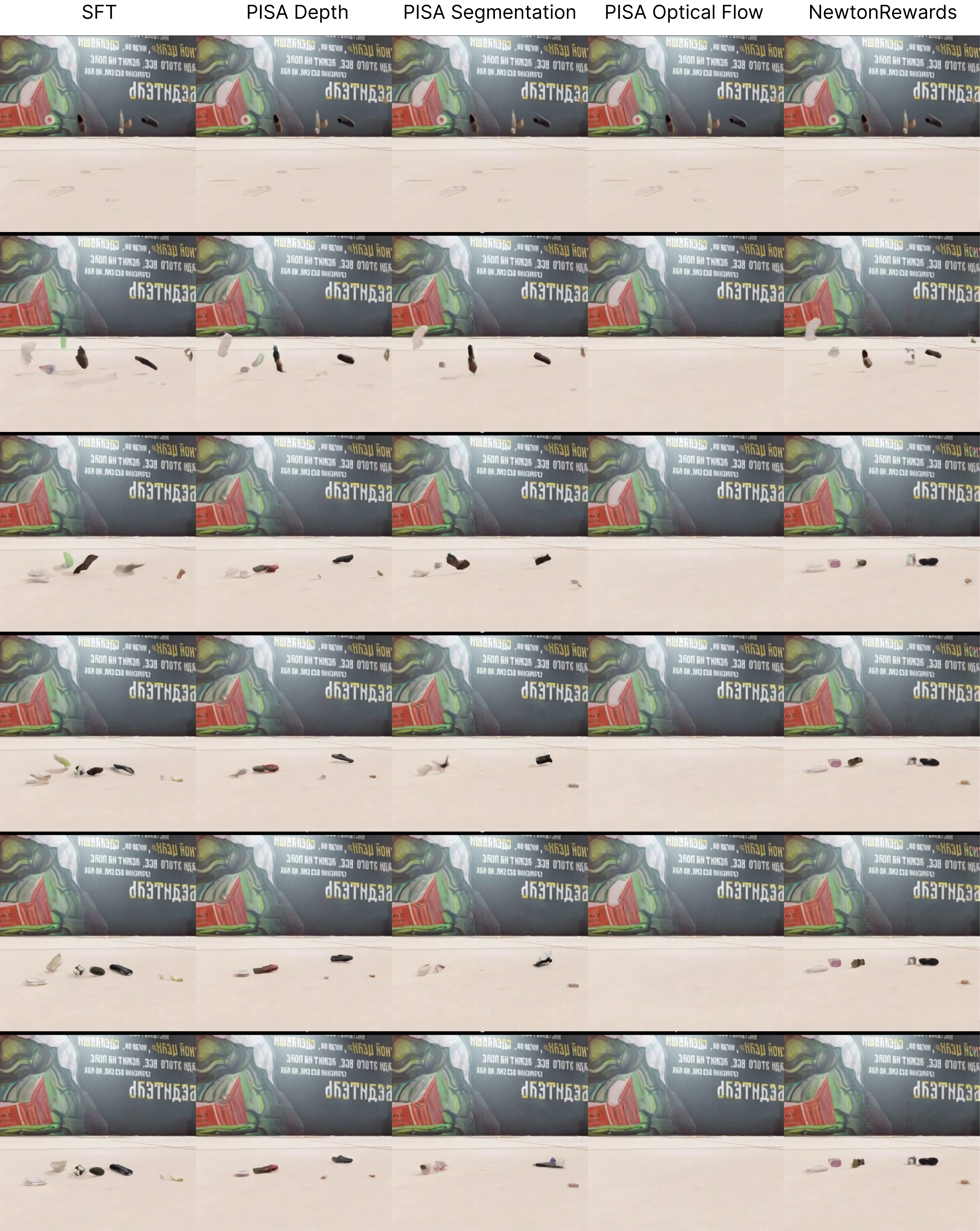}
\vspace{-3mm}
\caption{\small
\textbf{Qualitative comparison of post-training strategies on the \texttt{NewtonBench-60K} free fall scenario.} 
Clear differences appear when examining the temporal evolution across frames (top$\rightarrow$bottom). Under SFT and all PISA variants (Depth, Segmentation, Optical Flow), the objects frequently display inconsistent vertical acceleration and unstable trajectories sequence frames where items jitter, deviate laterally, or momentarily ``hover''. PISA Optical Flow particularly exhibits frame-to-frame jitter and irregular descent.
In contrast, NewtonRewards produces smooth, stable free-fall motion: objects drop along physically plausible vertical paths with consistent acceleration and minimal horizontal drift, closely matching true gravitational dynamics.
}
\label{fig:supp_frames_F}
\vspace{-10mm}
\end{figure*}

\begin{figure*}
\centering
\includegraphics[width=0.8\linewidth]{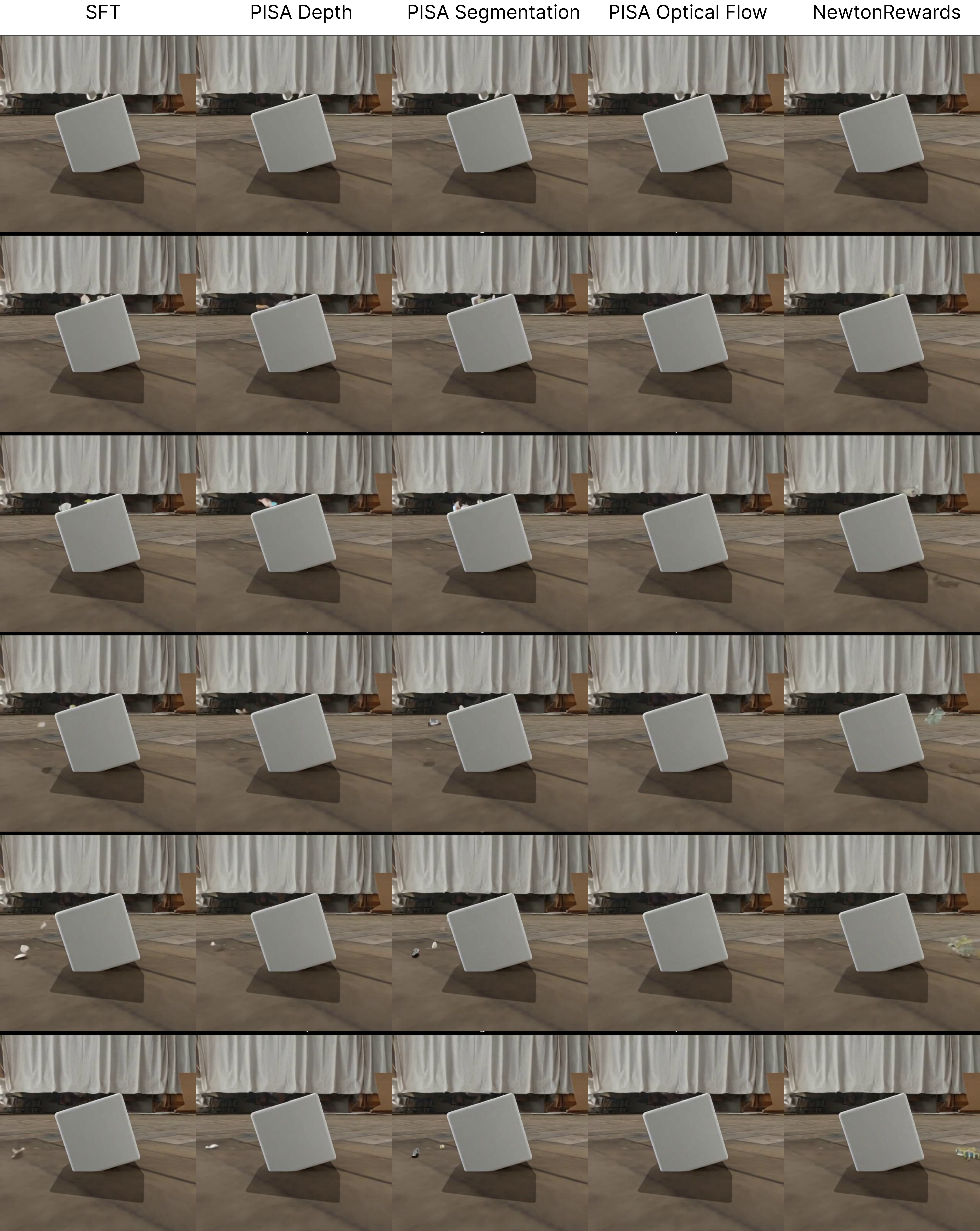}
\caption{\small
\textbf{Qualitative comparison of post-training strategies on the \texttt{NewtonBench-60K} ramp slide up scenario.}
Across temporal progression (top$\rightarrow$bottom), SFT and all PISA variants (Depth, Segmentation, Optical Flow) exhibit inconsistent contact dynamics. 
In SFT, PISA Depth, and PISA Segmentation, the objects are sliding oppositionally down; and in PISA Optical Flow the object just disappears.
In contrast, NewtonRewards produces coherent, physically grounded motion: the objects sliding up smoothly, with realistic frictional motion, and no frame-to-frame jitter. 
The resulting trajectory aligns closely with expected dynamics under gravity and surface friction. 
}
\label{fig:supp_frames_RU}
\vspace{-0.1cm}
\end{figure*}

\begin{figure*}
\centering
\includegraphics[width=0.8\linewidth]{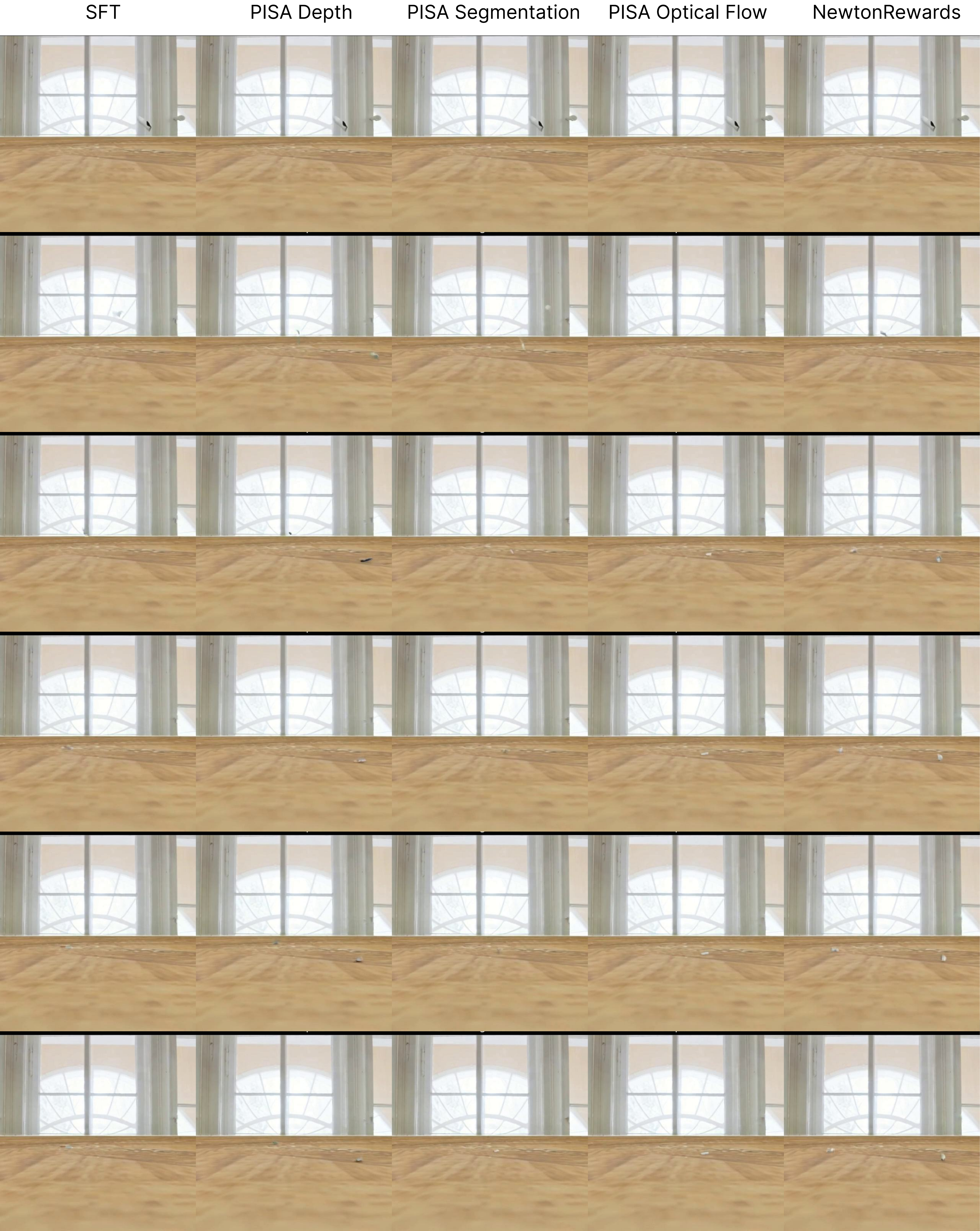}
\caption{\small
\textbf{Qualitative comparison of post-training strategies on the \texttt{NewtonBench-60K} horizontal throw scenario.}
Examining the temporal rollout (top$\rightarrow$bottom), SFT and all PISA variants (Depth, Segmentation, Optical Flow) exhibit inconsistent motion: objects either lose horizontal velocity too quickly, drift irregularly, or jitter frame-to-frame. Several PISA variants show abrupt slowdowns or curved, non-ballistic paths.
In contrast, NewtonRewards produces smooth, coherent trajectories that follow a realistic horizontal-throw profile: constant horizontal velocity, stable parabolic descent, and no unnatural jitter. 
The motion aligns closely with classical projectile dynamics, demonstrating significantly improved physical fidelity.
}
\label{fig:supp_frames_TH}
\vspace{-0.1cm}
\end{figure*}

\begin{figure*}
\centering
\includegraphics[width=0.8\linewidth]{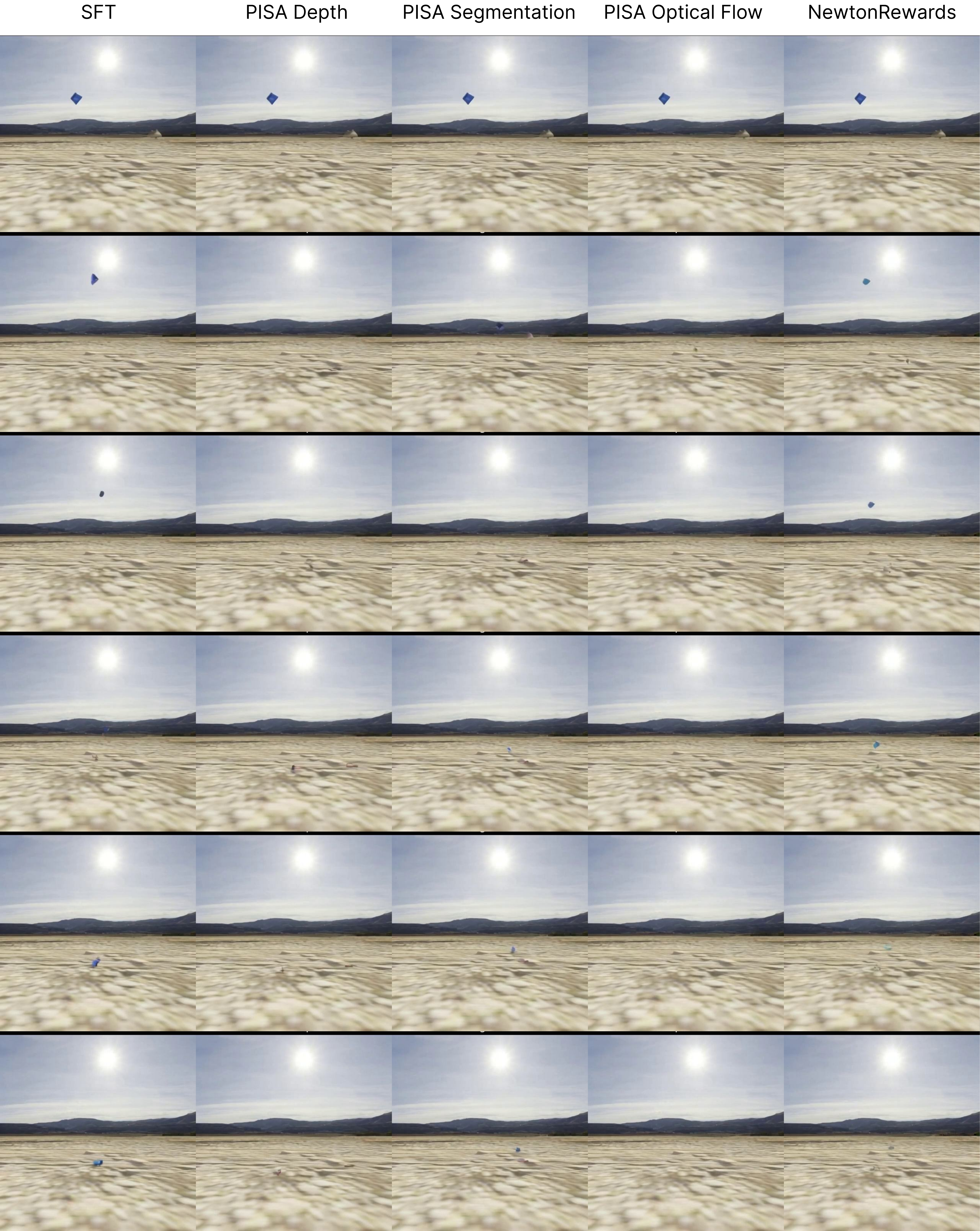}
\caption{\small
\textbf{Qualitative comparison of post-training strategies on the \texttt{NewtonBench-60K} parabolic throw scenario.}
Across the temporal rollout (top$\rightarrow$bottom), SFT and all PISA variants (Depth, Segmentation, Optical Flow) struggle to reproduce coherent parabolic motion. 
The thrown object exhibits noticeable inconsistencies--trajectory, abrupt velocity changes, or overly flattened arcs. 
In particular, objects disappear for the PISA Optical Flow case.
In contrast, NewtonRewards generates smooth, physically realistic motion: the object follows a stable parabolic path with consistent horizontal velocity and gravitationally governed vertical acceleration. 
}
\label{fig:supp_frames_TP}
\vspace{-0.1cm}
\end{figure*}

In addition to \cref{fig:qualitative}, we provide visual comparison in \cref{fig:supp_frames_F,fig:supp_frames_RU,fig:supp_frames_TH,fig:supp_frames_TP} between post-training strategies on our \texttt{NewtonBench-60K} for the remaining NMPs.

